\documentclass{article}
\usepackage{PRIMEarxiv}
\usepackage{authblk}
\usepackage[utf8]{inputenc} 
\usepackage[T1]{fontenc}    
\usepackage{hyperref}       
\usepackage{url}            
\usepackage{booktabs}       
\usepackage{amsfonts}       
\usepackage{nicefrac}       
\usepackage{microtype}      
\usepackage{lipsum}
\usepackage{fancyhdr}       
\usepackage{amsmath}
\usepackage{graphicx}       
\usepackage{algpseudocode}
\usepackage{algorithm}
\usepackage{enumitem}
\graphicspath{{media/}}     

\title{Memory-Augmented Generative Adversarial Transformers}

\author[1,2]{\bf Stephan Raaijmakers}
\author[2]{\bf Roos Bakker}
\author[3]{\bf Anita Cremers}
\author[4]{\bf Roy de Kleijn}
\author[5]{\bf Tom Kouwenhoven}
\author[5]{\bf Tessa Verhoef}
\affil[1]{Leiden University Centre for Linguistics (LUCL)}
\affil[2]{TNO, The Netherlands}
\affil[3]{University of Applied Sciences, Utrecht}
\affil[4]{Institute of Psychology, Leiden University}
\affil[5]{Leiden Institute of Advanced Computer Science (LIACS)}

\begin{document}
\maketitle

\begin{abstract}
Conversational AI systems that rely on Large Language Models, like Transformers, have difficulty interweaving external data (like facts) with the language they generate. Vanilla Transformer architectures are not designed for answering factual questions with high accuracy. This paper investigates a possible route for addressing this problem. We propose to extend the standard Transformer architecture with an additional memory bank holding extra information (such as facts drawn from a knowledge base), and an extra attention layer for addressing this memory. We add this augmented memory to a Generative Adversarial Network-inspired Transformer architecture. This setup allows for implementing arbitrary felicity conditions on the generated language of the Transformer. We first demonstrate how this machinery can be deployed for handling factual questions in goal-oriented dialogues. Secondly, we demonstrate that our approach can be useful for applications like {\it style adaptation} as well: the adaptation of utterances according to certain stylistic (external) constraints, like social properties of human interlocutors in dialogues.
\end{abstract}


\section{Introduction}

Transformers (\cite{vaswani2017attention}) are capable of producing natural, well-formed language with high degrees of fluency. They are responsible for the latest commercial and open source large language models (LLMs) like BLOOM (\cite{workshop2023bloom}) and the GPT-models powering ChatGPT.  Transformer architectures produce such language models either through a full encoder-decoder combination (e.g. T5, \cite{RaffelSRLNMZLL20}), just the encoder (BERT, \cite{devlin-etal-2019-bert}) or the decoder (e.g. the GPT models, \cite{ouyang2022training}). In general, probabilistic language models decompose the probability of word sequences $w_1,\ldots, w_n$ into a product of conditional probabilities, estimated from raw textual data:

\begin{equation}
P(w_t\mid w_1\ldots w_{t-1})= \displaystyle\prod_{t=1}^n P(w_t\mid w_{<t})
\end{equation}

The size of neural network-driven Large Language Models is determined by the amount of training data, the amount of model parameters, and the amount of GPU flops for training the models. LLMs treat words as points in a high-dimensional vector space ({\it encoding}) and complete (encoded) word sequences with generated, subsequent words ({\it decoding}). Such probabilistic neural models can be summarized succinctly for a left-to-right situation (generating words on the basis of left context) as
\begin{equation}
P(w_t\mid w_1\ldots w_{t-1})=\exp{(Emb^\top_{w_t} f_\theta(w_1\ldots w_{t-1}))} 
\end{equation}
with $Emb^\top_{w_t}$ the embedding of token $w_t$, and $f_\theta(\cdot)$ a parameterized function that, based on the learned parameter settings $\theta$ of the neural network architecture used, encodes the preceding word sequence $w_1\ldots w_{t-1}$ into a combined vector representation.  A more general form is
\begin{equation}
P(w_t\mid w_1\ldots w_{t-1})= \displaystyle\prod_{t=1}^n P_{\theta}(w_t\mid w_{<t})
\end{equation}
LLMs minimize during training the negative log-likelihood over a training corpus (a collection of documents $D$, $D[d]$ being the $d$-th document in $D$, and $|D[d]|$ its document length in terms of tokens):

\begin{equation}
\mathcal{L}(D)=-\displaystyle\sum_{d=1}^{\mid D\mid}\sum_{t=1}^{\mid D[d]\mid} \log p_{\theta}(w_t^{D[d]}\mid w_{<t}^{D[d]}).
\end{equation}

In conversational AI, and specifically in goal-oriented dialogues, additional conditions on generated utterances typically apply. In such dialogues, conversational agents need to respond to users with appropriate utterances that are not only linguistically correct but also provide correct information, and are stylistically acceptable in the current dialogue context (e.g. \cite{liebrecht2021}). 
Unfortunately, vanilla Transformers are not equipped for such tasks. Their purpose is to generate language from language, conditioned on attention patterns ({\it self-attention} and {\it intra-attention}). A Transformer typically learns to convert input sequences into output sequences, encoding their input with attention values, and generating output from the encoded representations. Standard Transformers cannot readily produce factual information in response to user input, e.g. for answering questions (\cite{lee_etal_2022}): all knowledge they contain is derived from their underlying language model. In addition,  they appear to be prone to {\it lexical hallucinations} (\cite{ziweietal23}), generating random output words based on small, unexpected perturbations in their input data. These innate deficiencies make them unsuitable for accurate question answering. This paper addresses the question of how to improve this situation for conversational systems based on Transformer-produced LLMs ("conversational LLMs"). Augmenting the memory of Transformers with external (e.g. factual) data is a well-known tactic (see Section \ref{relatedwork} below). We argue for adding another control facility to such memory-augmented Transformers: a generative adversarial training tactic that allows for exerting additional conditions (such as factual compliance) on the utterances generated by Transformers. We will first discuss such an adversarial control mechanism, and then present the combination of that mechanism with a memory augmented Transformer architecture.  

\section{Generative Adversarial Transformers}
\label{sec:headings}
A generative adversarial network (GAN, \cite{goodfellow2014generative}) is an implementation of a {\it zero-sum} game, where two parties interact with each other: a {\it discriminator} and a {\it generator}. In zero-sum games, the loss of one party benefits the other party. Zero-sum games are commonly expressed in terms of a {\it value function} $V(D,G)$ which, for a discriminator $D$ and a generator $G$, expresses the quantities to be minimized and maximized during training. 
In particular, the discriminator $D$ seeks to {\it maximize} the  objective of discriminating between a reference data distribution and the 'fake' data generated by the generator, which initially will be noisy, but will gradually start approximating the true data distribution. The generator attempts to {\it minimize} the dispersion of its own data distribution and the reference data. In this context, zero-sum implies that the amount to which the discriminator discerns successfully between fake and real data points translates to a penalty for the generator: the better the discriminator performance (i.e. the lower its error when discriminating between true and faked data), the more severe the penalty for the generator: it must work harder to mislead the discriminator, and thereby increase the discriminator error. Conditional GANs are a specific type of networks that are conditioned on desired output labels (\cite{mirza2014conditional}).
We propose in this paper a type of conditional Generative Adversarial Transformer (GAT). Both generator and discriminator are implemented as Transformers, trained through the adversarial zero-sum game of GANs.  The generator can be conditioned on additional loss functions $\cal{L}_G$ and external data ${\cal M}$. The GAT has the following constrained value function $V$:
\begin{equation}
\underset{D}{\operatorname{min}}\ \underset{G}{\operatorname{max}}\ V(D,G)=\mathbb{E}_{x\sim p(x}\left[\log(D(x,m_x)\right] + \mathbb{E}_{z\sim p_z(z)}\left[\log(1-D(G(z, m_z, \lambda^*))\right]
\end{equation}
with $p_x$ the true data distribution, $p_z$ the noisy distribution from the generator, and
\begin{equation}
\lambda^*=\lambda_1^n\in\mathcal{L}_G=\lambda_1(\cdot)+\ldots+\lambda_n(\cdot).
\end{equation}
with the default that $\lambda^*$ is a zero function $f(\cdot)=0$.

$D$ and $G$ are parameterized functions ($m_x$ the external data associated with data point $x$):  
\begin{equation}
\begin{array}{c}
D(x, m_x;\theta_D)\\[.7cm]
G(x, m_x, \lambda^*;\theta_G)
\end{array}
\end{equation}
The expected values are defined as:
\begin{equation}
\mathbb{E}_{x\sim p(x)}\left[\log(D(x,m_x)\right]= \int \log(D(x,m_x)) p(x) dx\\[.7cm]
\end{equation}
and
\begin{equation}
\mathbb{E}_{z\sim p_z(z)}\left[\log(1-D(G(z,m_z,\lambda^*)),m_z)\right] = \int \log(1-D(G(z,m_z,\lambda^*),m_z)) p(z) dz
\end{equation}

This expresses that, for a given data point $x$,  both the generator $G$ and the discriminator $D$ take the corresponding external data $m_x$ into account, and that the generator may be subjected to additive, generator-specific loss functions. According to this design, a GAT can be conditioned in three manners: (1) through the external data that accompanies the normal stream of input data; (2) through additional felicity conditions imposed on the generator; (3) by combining (1) and (2). This opens up possibilities for checking factual adherence of these models. Practically, we implement such conditions as loss functions on the generator, and we add their results through summation to the default Transformer loss function scores (sparse categorical cross entropy, measuring overlap between predicted utterance and ground truth utterance). 
In terms of language modeling, this means we redefine negative log-likelihood as follows:
\begin{equation}
\mathcal{L}(D)=-\displaystyle\sum_{d=1}^{\mid D\mid}\sum_{t=1}^{\mid D[d]\mid} \log p_{\theta}(w_t^{D[d]}\mid w_{<t}^{D[d]},m_d, \mathcal{L}_G).
\end{equation}
with $m_d$ the external data linked to "document" $d$ (a training data item).

Next, we need to make the extra, external data available to the conditional Generative Adversarial Transformer through {\it memory-augmentation}.

\subsection{Memory-augmented Conditional Generative Adversarial Transformers}
\label{sec:others}
\begin{figure}[htb]
\begin{center}
\includegraphics[scale=0.15]{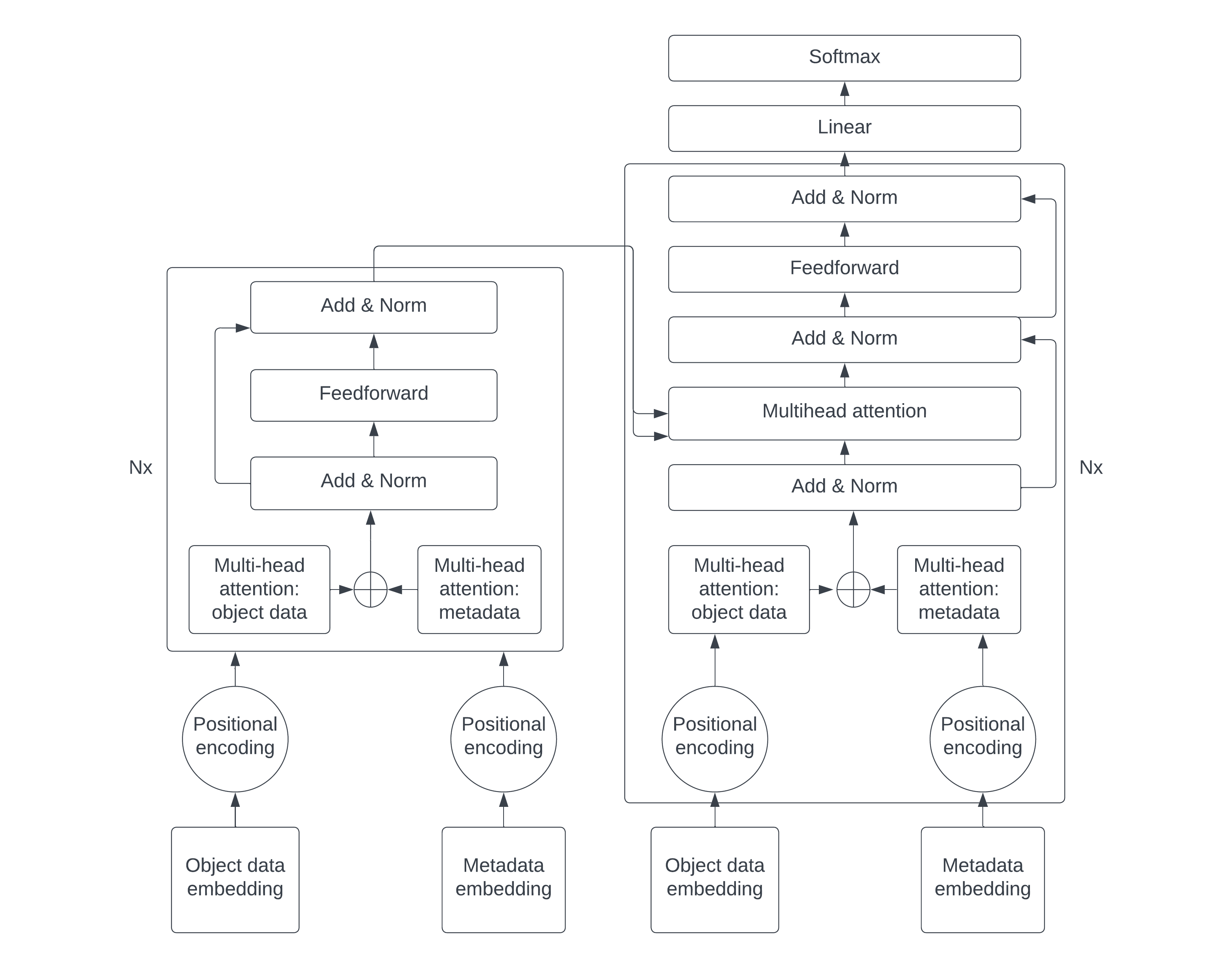}
\end{center}
\caption{The memory-augmented Transformer with its encoder (left block) and decoder (right block).}
\label{enc-dec-GAT}
\end{figure}

\begin{figure}[htb]
\begin{center}
\includegraphics[scale=0.2]{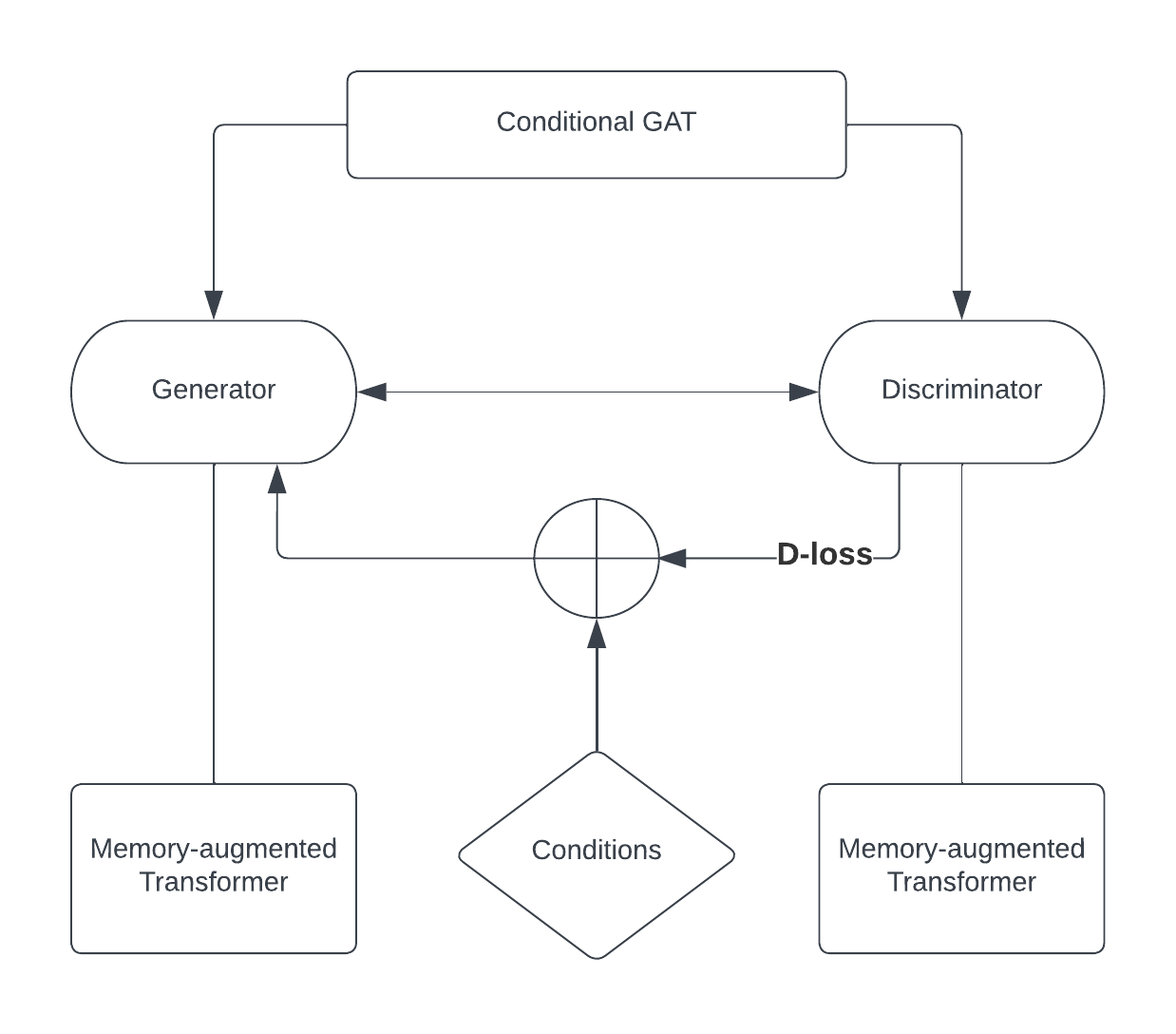}
\end{center}
\caption{The conditional Generative Adversarial Transformer (GAT), built up from two memory-augmented Transformers. The generator is equipped with additional loss functions conditioning its output. Notice how the generator receives summed losses from the additional loss functions ("Conditions") and the discriminator ("D-loss").}
\label{GAT-Controller}
\end{figure}

 We equip the GAT generator and discriminator with an additional attention layer addressing a separate stream of external data, creating a {\it memory-augmented} Transformer. The external data is aligned with the original input data on an item-by-item basis, and can be optionally empty or even address  data that is part of the original input, as a means of emphasis. Specifically, in both encoder and decoder components of the Transformer, we add an additional multihead attention layer for external data.  Every head in this layer computes attention using the standard Transformer {\it Query/Key/Value} mechanism (\cite{vaswani2017attention}). For handling external data, we let {\it Query}=inputs, {\it Key}=external data and {\it Value}=external data. Attention for both original input data and external data is summed in both the encoder and decoder blocks of the Transformer, similar to the approach of \cite{GhazvininejadBC18}. Figure \ref{enc-dec-GAT} illustrates the encoder and decoder blocks of the memory-augmented Transformer. The controlling Generative Adversarial Transformer is illustrated in Figure \ref{GAT-Controller}

As discussed, the generators in GANs generate initially noisy data that becomes scrutinized or rewarded by the discriminators. The worse the discriminator discerns the generated data from real data, the higher the reward for the generator. In the case of text-based Transformers,  the output of the generators, per training epoch, is treated as noisy data, all of which is labeled as 'fake'. The generator can be conditioned with arbitrary loss functions, measuring the quality of the generated answers given the corresponding questions. These loss functions are supplementary to the normal loss function imposed on the Transformer, which measures the dispersion of the generated answers with respect to the ground truth training data using sparse categorical entropy over the integer-encoded words in inputs and predictions. In subsection \ref{expcond}, we outline a sample loss function used in our experiments.

\section{Related work}\label{relatedwork}
There are different perspectives on conditioning conversational Large Language Models on external data. Our work can be related to - and contrasted with- five major types of approaches. 
\\[.3cm]
{\bf Black-box LLMs and fine-tuning} One obvious bottleneck for fine-tuning and memory augmentation approaches to conditional language generation is the necessity to invest in (re)training LLMs. Recent approaches like \cite{peng2023check}  attempt to circumvent this restriction by connecting pre-trained large language models  ("black-box LLMs") to external data sources directly. This differs from our approach, where we effectively train LLMs from the ground up. Alternatively, many approaches focus on {\it fine-tuning} pre-trained Large Language Models for conditional conversation generation. An early example is \cite{GhazvininejadBC18}, who propose to combine conversational and non-conversational, factual data with Memory Networks through multi-task learning, where conversational and factual text generations tasks are learned in conjunction. While their approach differs from the approach we outline in this paper from an architectural point of view, it is related in combining encoded factual and non-factual information into one representation, as we will point out below.  
\\[.3cm]
{\bf Memory and retrieval augmentation} Adding external memory banks to LLMs is known as {\it memory augmentation}. In \cite{KhandelwalLJZL20}, Large Language Models are interpolated with {\it k}-nearest neighbor ({\it k}-NN) models, using textual similarity of current LLM context with retrieved context(and their completions) for generating interpolated completions. Since {\it k}-NN models are memory-based models that store exceptions very well (\cite{DaelemansBZ99}, nearest neighbor LLMs can handle rare patterns and factual information.  In a similar vein, {\it memory augmentation} approaches aim to augment the memory capacities of the neural architectures that produce Large Language Models, by adding extra memory facilities that address external information, like facts from a knowledge base, e.g. \cite{tacl_a_00476}, where Large Language Models are conditioned on structured external information organized in knowledge graphs.  Retrieval-Augmented Generation (RAG, \cite{lewis2020}) aims to connect an LLM to external data sources, by mapping user queries or prompts to vectorized representations used as queries for external data sources (like databases). Retrieval results
are handled by the LLM to formulate replies that depend on both the user prompt data and the retrieval result. Unlike RAG, we attempt to internalize such external knowledge in the LLM itself.
In \cite{fevry2020entities}, a separate entity memory ("Entities as Experts") is linked to a Transformer model, that is used for filling in contextual entity mentionings with learned entity embeddings, using contextual similarity matching. Our approach omits multi-step involvement of entity data - we do not explicitly encode separate entity information, but rather force Transformers to focus on external information (not restricted to entities) through the native attention mechanism of Transformers. In \cite{verga-etal-2021-adaptable}, the Entities as Experts approach is extended with additional knowledge base facts pertaining to (again) entities. The work of \cite{zhong2022training} augments the Transformer memory with short term token (i.e. lexical) memory, long term token memory and external token memory.  Their approach differs from ours in the type of memory data (lexical tokens), and the absence of additional constraint facilities. In contrast, our approach is agnostic to the type of information in the external data memory bank, and allows for arbitrary constraints on outputs. In \cite{dejong2022mention}, a separate entity-based external model is proposed, the entirety to which the Transformer pays attention during utterance generation. Our work relates to this approach, in proposing a memory bank for external information, but is more general in allowing for any type of information, including emphasizing existing, internal information from the original input memory, and in allowing for the expression of felicity conditions on the Transformer that can address the extra data memory and the produced responses. Finally, the recent models of \cite{shinn2023} propose verbal feedback memory buffers for self-reflection by LLMs.
\\[.3cm]
{\bf Prompting} \cite{ram2023incontext} and \cite{shi2023replug} propose to leverage {\it prompting} (or {\it in-context learning}, \cite{min2022rethinking, zhou2023large}) facilities of current Large Language Models: inserting factual information into a short-term memory buffer of an existing Large Language Model, forcing it to take such information in account when generating new utterances. Our work differs from this work in not explicitly mixing input data with controlling external data, but instead keeping external apart in a separate memory buffer, and by allowing for explicit conditions on outputs through the adversarial organization of the Transformer model.
\\[.3cm]
{\bf Generative Adversarial Transformers}. Finally, in the image analysis field, Generative Adversarial Transformer models have been proposed, e.g. by \cite{hudson2022generative}.
In \cite{zeng2020style}, a Generative Adversarial Transformer architecture has been proposed that exclusively focuses on textual style transfer. Our work can be seen as a generalization of this idea, in being agnostic to the task at hand. 

\section{Experiments}
In a number of experiments we assess the benefits of the external data memory and conditioning the generator of the adversarial Transformers. First, we describe our data, and subsequently we outline our experimental setup.  

\subsection{Data}
In our experiments, we use two datasets:
\begin{itemize}
\item The data from (\cite{EricKCM17}), referred to below as CAR. This is multi-turn conversational data gathered through a {\it Wizard-of-Oz} experiment for the automotive domain, describing goal-oriented dialogues between humans and a car navigation system. 
\item The Personalized bAbI data (\cite{joshi2017personalization}). This dataset builds upon the well-known bAbI dialogue datasets (\cite{bordes2017learning}, and consists of goal-oriented (restaurant table booking) dialogues extended with gender and age information about the human interlocutor. The overarching task is to train dialogue systems to adapt their style to human interlocutor characteristics. 
\end{itemize}
These datasets are described in more detail below.

\subsubsection*{Factual Question-Answering: CAR\label{emexp}}
The CAR data contains interactions of humans with a (fictive, i.e. human-imitated) car voice assistant. Drivers can ask the assistant about weather conditions, nearby points of interests ("poi's"), and express navigation instructions. The data is grounded with a knowledge base containing destinations, addresses, weather information and points of interest (POIs). It consists of 3,031 dialogues in  three different domain types, all dealing with in-car personal assistants. Uniquely sorting the preprocessed data for the three stages resulted in 5,958 single turn cases for stage 1, 3,864 single turn cases for stage 2, and 3,041 single turn cases for stage 3. Examples are listed in Table \ref{data-example-EM}.

\begin{table}[H]
\begin{center}
\begin{tabular}{p{7cm}|p{7cm}}
\toprule
{\bf Question/Instruction} & {\bf Response}\\ 
\midrule
Where's the nearest parking garage?	& The nearest parking garage is Dish Parking at 550 Alester Ave. Would you like directions there?\\[.3cm]
Yes, please set directions via a route that avoids all heavy traffic if possible. & It looks like there is a road block being reported on the route but I will still find the quickest route to 550 Alester Ave. \\[.3cm]
Thanks so much for your help. & You're very welcome!\\[.3cm]
Show me the closest location where i can get chinese food. & The closest chinese restaurant is PF Changs, located 5 miles away.
\\ 
\bottomrule
\end{tabular}
\\[.7cm]
\caption{Examples from the data from \cite{EricKCM17}.}
\label{data-example-EM}
\end{center}
\end{table}

In order to obtain fine-grained control over external data for answer generation, we carve up this dialogue process into three stages:
\begin{enumerate}
\item[Stage 1:] Slot detection.  Using the knowledge base (KB) information associated with the dialogues in this dataset, we reverse engineer  a slot-tagged version of the dataset, mapping words and phrases to the associated slot names in the KB information that accompanies each turn in a dialogue. We train an adversarial Transformer on the detection of entity slots (names, locations, distances, etc.) from raw data. This Transformer has an empty external data memory bank, since external data do not play a role here. The output of this stage consists of entity-attenuated utterances (Table \ref{data-example-EM})\footnote{The reverse data engineering process was not fault-proof and led in a number of cases to ill-formed data (missing entities, or mistagged entities).}.  
\item[Stage 2:] Slot mapping. We train a memory-augmented adversarial Transformer on mapping entity-attenuated source utterances (questions, instructions) to entity-attenuated target utterances (responses). In this stage, external data consists of the entity slots specified in the source utterances, which serves to  {\it emphasize} this information for response generation. 
\item[Stage 3:] Slot filling. We train a memory-augmented adversarial Transformer on mapping the entity-attenuated responses from stage 2 to instantiated utterances, with actual values substituted for slot names.  In this stage, external data consists of factual database values.
\end{enumerate}

\begin{figure}[htb]
\begin{center}
\includegraphics[scale=0.2]{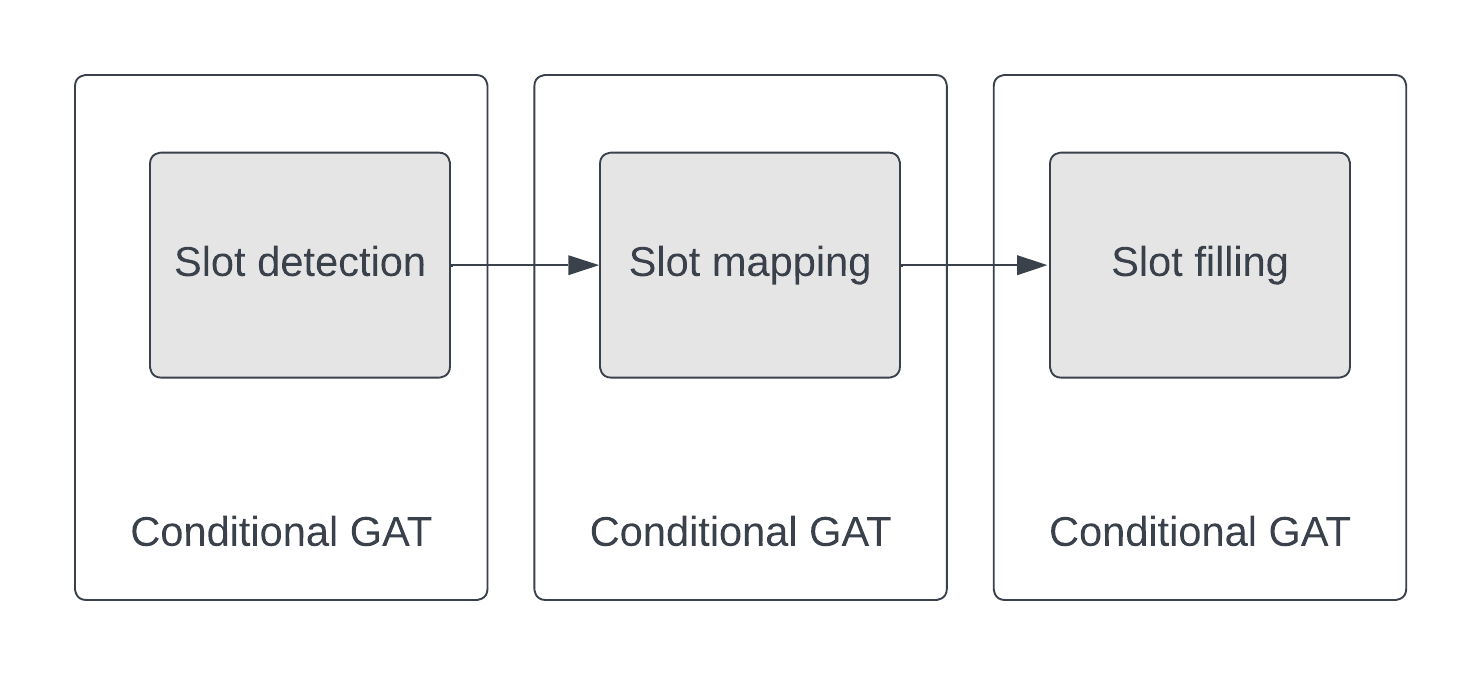}
\end{center}
\caption{Setup for factual question answering. Three separate memory-augmented Transformers, each trained with their own conditional GAT, address different stages in the process.}
\label{EM-setup}
\end{figure}

This setup is illustrated in Figure \ref{EM-setup}. Sample processed data from the CAR dataset is listed in Table \ref{data-example-EM-proc}. 

\begin{table}[h]
\begin{tabular}{p{5cm}|p{5cm}|p{5cm}}
\toprule
Stage 1: slot detection  &  Stage 2: slot mapping & Stage 3: slot filling\\ 
\midrule
Input: can you find me the fastest route to a parking garage & Input: can you find me the poidistance route to a poitype & Input: poi is poidistance away\\[.7cm]
External data: none & External data: poitype poidistance &  External data: poitype:parking garage poi:webster garage poidistance:4 miles\\ [.3cm]
Output: can you find me the poidistance route to a poitype & Output: poi is poidistance away & Output: webster garage is 4 miles away
\\ 
\bottomrule
\end{tabular}
\\[.7cm]
\caption{Examples from our attenuated version of the data from \cite{EricKCM17}, across the three stages.}
\label{data-example-EM-proc}
\end{table}

In order to produce the external data for stage 2 and 3 (which merges this data with the template generated by stage 2), stage 1 delivers crucial information that needs to be converted into a database query. By aligning the entities assigned to words in stage 0 (in the example in Table \ref{data-example-EM}: {\it fastest} $\mapsto$ {\it poidistance}; {\it parking garage}$\mapsto$ {\it poitype}), we can construct a database query with restrictions, to be resolved against a knowledge base. 
For instance, for the listed example, such a query submitted to a retrieval engine like Elasticsearch\footnote{https://www.elastic.co} could look like a {\tt GET} operation (Figure \ref{elastic-query}).
\begin{figure}[h!]
\begin{verbatim}
GET /_search
{
  "query": { 
      "filter": [ 
        { "poitype": "parking garage" },
        { "poidistance": { "fastest": { "current-location": "..." }}}
      ]
  }
}
\end{verbatim}
\caption{Sample Elastic query.}
\label{elastic-query}
\end{figure}
The result of this query should be the external data for stage 2, in our example\footnote{In the CAR data we used, such queries have been resolved already.}
\begin{itemize}
\item \tt{poitype:parking garage, poi:webster garage poidistance:4 miles}.
\end{itemize}

The external data for stage 2 is essentially a repetition of the extracted slots in the input buffer, produced by stage 1. By representing these slots without context, explicitly in a separate memory buffer and subjecting them to separate attention, we instruct  the Transformer to focus on this information during response generation.  In summary, for the CAR data we split up dialogue utterance generation into three separate generative steps: an initial slot extraction step, generating attenuated utterances with entities (like dates, destinations, locations) replaced by slot names; a subsequent answer generation step that maps slot-attenuated utterances (questions, commands) to slot-attenuated responses, and a final step that fills in attenuated slots with values coming from a knowledge base. Under this perspective --which can trivially be generalized to other datasets--  question-answering becomes a slot extraction and slot filling problem.  In our current experiments, we did not train the three Conditional GAT models in an end-to-end fashion, they were trained independently.

\subsubsection*{Style adaptation: Personalized bAbI}
For the Personalized bAbI data, we focus on different external data: the gender and age information associated with the human interlocutor. The data contains 5 tasks, in different sizes (large and small). We used the dataset for task 3, a small dataset comprising 1,369 single question/answer turns \footnote{See https://github.com/chaitjo/personalized-dialog for further details of this data.}.The human-provided initial question and the age/gender information about the interlocutor serve as input for the answer generation process. Typically, age and gender are reflected in the style of the response (higher age leading to more formal, polite answers, for instance). Some examples are listed in Table \ref{data-pb}.
\begin{table}[h!]
\begin{tabular}{p{5cm}|p{5cm}|p{5cm}}
\toprule
Question & External data & Answer\\ 
\midrule
bombay please  & female elderly&  would you mind telling me how many guests shall be at your table\\[.3cm]
bombay please  & female young&    how many are you\\[.3cm]
bombay please  & male middle-aged&        ok sir i'm looking for options for you\\[.3cm]
can you book a table & female elderly &  thank you madam i shall start the reservation now\\
\bottomrule
\end{tabular}
\\[.7cm]
\caption{Examples from the Personalized bAbI data from \cite{joshi2017personalization}.}
\label{data-pb}
\end{table}
This data is processed in one step: our Transformer maps input and external data directly to output.

\subsection{Experimental conditions\label{expcond}}
We carried out three experiments for our two datasets, listed in Table \ref{experiments}.
\begin{table}[H]
\begin{center}
\begin{tabular}{llp{10cm}}
\toprule
Experiment & Data &Description \\
\midrule
EXP1 & CAR data & For stages 1 and 3 for the CAR setup (Table \ref{data-example-EM-proc}), this experiment verifies the linguistic capability of the memory-augmented Transformers. Stage 1 does not use external data, and stage 3 uses external data per definition, to fill in database values in slots detected by stage 1. \\[.3cm]
EXP2 & CAR data & This experiment investigates the use of external data for stage 2 of the CAR setup and the benefits of the POI loss function as a condition on the generator. Total loss is computed by summing standard loss and POI loss, and performance is compared to memory-augmented Transformers with emptied external data.\\[.3cm]
EXP3 & Personalized bAbI data & The benefit of accessing gender and age-related external data. Performance is compared to memory-augmented Transformers with emptied external data. No specific loss function is imposed on the generator.\\
\bottomrule
\end{tabular}
\\[.3cm]
\caption{Our experiments.}
\label{experiments}
\end{center}
\end{table}

As mentioned, for the CAR dataset, stage 1 does not use external data, and stage 3 uses external data per definition for filling in slots with values. We therefore only report on the effect of external data or no external data for stage 2 in our experiments. Notice that the three models for stages 1-3 were not trained end-to-end; they were trained independently (see Section \ref{futurework}). Additionally, we investigate the effect of a loss function that specifically addresses the overlap in external data between query and generated answer: the {\it point of interest (POI)} loss function ($\lambda_{poi}$), which we define as a function based on recall and precision for POI slot names:
\begin{equation}
\begin{array}{l}
Precision^{poi}=\frac{TP^{poi}}{TP^{poi}+FP^{poi}}\\[.3cm]
Recall^{poi}=\frac{TP^{poi}}{TP^{poi}+FN^{poi}}\\[.3cm]
F_1^{poi}=2\cdot\frac{Precision^{poi}Recall^{poi}}{Precision^{poi}+Recall^{poi}}\\[.3cm]
\lambda_{poi}=1-F_1^{poi}
\end{array}
\end{equation}

This loss function measures the amount to which POI slot names in the extra data memory appear in the generated predictions. It only checks the presence of slot names, not specific values.

As for the Personalized bAbI data, we test for the benefit of exposing Transformers to the gender and age-related external data
(see Table \ref{data-pb}).

In our experiments, we uniformly split off a randomized, fixed (over runs) portion of 20\% of training data for testing purposes and trained on the remaining 80\% of data\footnote{The Transformers parameters were: batch size=8, embedding dimension for input (words) and external data=256, number of attention heads=8.}.
All Transformers were trained for 1,000 epochs on a single Tesla T4 16GB GPU. We measure results with the {\tt sacrebleu} (\cite{sacrebleu}) toolkit\footnote{We ran {\tt sacrebleu} with the signature 
{\tt -l en-en -m bleu chrf ter --chrf-word-order 4 -b  -w 4 --paired-bs}.}.
This toolkit computes a cumulative sentence BLEU scores based on word 4-grams (BLEU-4), and additionally computes chrF2 (a character n-gram-based F-score for the match between two utterances, \cite{popovic-2017-chrf}) and TER (Translation Edit Error, \cite{snover-etal-2006-study}), which measures the amount of edit steps needed to convert a machine-generated utterance  into a reference utterance (lower TER is better). The {\tt sacrebleu} toolkit also compares results across multiple conditions, with one of them being a baseline, to a reference dataset and reports eventual  statistical significance.
Additionally, we measure outcomes with the following set of metrics, using the {\tt nlg-eval} toolkit (see \cite{nlgeval} for details):
\begin{itemize}
\item METEOR: computes an F-score based on a fuzzy alignment of unigrams between source and target utterances, and has been shown to align better with human judgment (\cite{meteor}).
\item ROUGE-L: computes an F-score addressing the {\it longest common subsequence}  between source and target utterances.
\item Embedding similarities: SkipThoughtsCosineSimilarity (computes cosine similarity defined on skip-thought sentence embeddings, see \cite{kiros2015skip}), EmbeddingAverageCosineSimilarity (cosine similarity based on averaged word embeddings), VectorExtremaCosineSimilarity (computes cosine similarity between sentence embeddings using extreme values of the constituting word embeddings).
\item GreedyMatchingScore: computes cosine similarity based on different pairings of the words in source and target sentences.
\end{itemize}

\subsection{Results}
Below we present the results for our three experiments\footnote{The Appendix lists sample responses for the CAR and Personalized bAbI tasks.}. Note that BLEU scores in the range of 50-60 are generally seen as high (see e.g. \cite{google-bleu}).

{\bf EXP1 - Slot detection and filling for the CAR data}. Tables \ref{sb-em-s1-s3} and \ref{nlg-em-s1-s3} list results of stages 1 (slot detection) and 3 (slot filling) for the CAR dataset. 
\begin{table}[H]
\begin{center}
\begin{tabular}{rp{3cm}p{3cm}p{3cm}}
\toprule
 System & BLEU-4 ($\mu$ ± 95\% CI) & chrF2++++ ($\mu$ ±95\% CI) & TER ($\mu$ ± 95\% CI) \\
\midrule
 Stage 1 & 81.7 (81.6311 ± 1.8414)      &  86.5 (86.4899 ± 1.3697)  &   12.5 (12.5455 ± 1.3463) \\ \\
 Stage 3 & 52.0 (52.0076 ± 2.2496) &    61.6 (61.5831 ± 1.7601) &   32.5 (32.5215 ± 1.7969) \\
\bottomrule
\end{tabular}
\\[.3cm]
\caption{EXP1: the {\tt sacrebleu} results produced for the CAR data for stages 1 and 3.}
\label{sb-em-s1-s3}
\end{center}
\end{table}

\begin{table}[H]
\begin{center}
\begin{tabular}{r|p{3cm}|p{3cm}}
\toprule
 Metric & Stage 1 & Stage 3 \\
 \midrule
BLEU-1 & 90.5 & 71.6 \\ 
BLEU-2 & 87.5  & 64.4\\ 
BLEU-3 &84.8& 58.2\\ 
BLUE-4 & 82.1 & 52.5\\ 
METEOR & 56.4& 36.6\\ 
ROUGE-L & 92.7  &75\\ 
SkipThoughtsCosineSimilarity &91.9 & 78.3\\ 
EmbeddingAverageCosineSimilarity & 97.9 &94.2\\ 
VectorExtremaCosineSimilarity & 94.9  & 75.2\\ 
GreedyMatchingScore & 97.9  & 88.4\\ 
\bottomrule
\end{tabular}
\\[.3cm]
\caption{EXP1: metric results produced by {\tt nlg-eval} for the CAR data for stages 1 (slot extraction) and stage 3 (slot filling).}\label{nlg-em-s1-s3}
\end{center}
\end{table}

While displaying relatively strong BLEU (and other) scores, stage 3 results show room for improvement in terms of factual adherence.  Frequently, we witnessed factual hallucinations like:

{\it Input}: the poidistance is poi  poidistance away at poiaddress\\
{\it External data}: poidistance:3 miles poiaddress:9981 archuleta ave poitype:coffee or tea place poitrafficinfo:moderate traffic poi:peets coffee\\
{\it Ground truth}: the closest is peets coffee \underline {3 miles} away at 9981 archuleta ave\\
{\it Prediction}: the nearest is peets coffee \underline{1 miles} away at 9981 archuleta ave\\[.3cm]
It is clear we need to impose additional value-checking loss functions here, which should come as no surprise, since the POI loss function only checks for the restoration of slot names in the generated output, not their values. 

{\bf EXP2 - Slot mapping for the CAR data}. In Table \ref{em-sb-s2}, the {\tt sacrebleu} scores for stage 2 for the CAR dataset are displayed. First, we compare using external data with the standard loss to using external data where POI loss is added to standard loss. Second, we compare the basic setting of using no external data and just the standard loss to using external data and POI loss added to standard loss. Third, we compare the use of  external data plus the combined POI and standard loss to not using external data but still using the combined POI loss and standard loss. Here, the POI loss is based on the overlap of input buffer POIs with predicted POIs in the generated utterances. This latter comparison specifically assesses the importance of the memory buffer.

\begin{table}[H]
\begin{center}
\begin{tabular}{rp{3cm}p{3cm}p{3cm}}
\toprule
 System & BLEU-4 ($\mu$ ± 95\% CI) & chrF2++++ ($\mu$ ±95\% CI) & TER ($\mu$ ± 95\% CI) \\
\midrule
 External data, standard loss &         9.9 (9.9140 ± 1.0474)         &        29.9 (29.8409 ± 1.2403)        &       90.1 (90.1385 ± 2.3366)        \\
       External data, POI loss+standard loss& {\bf 11.2} (11.1627 ± 1.1571) (p = 0.0230)* & {\bf 31.2} (31.1684 ± 1.2127) (p = 0.0090)* & 90.8 (90.9146 ± 2.7216) (p = 0.2208) \\
\midrule
 No external data, standard loss&       10.3 (10.2779 ± 1.1844)        &        29.9 (29.8587 ± 1.2475)        &       89.2 (89.2030 ± 2.4749)        \\
          External data, POI loss+standard loss& 11.2 (11.1627 ± 1.1571) (p = 0.0819) & {\bf 31.2} (31.1684 ± 1.2127) (p = 0.0140)* & 90.8 (90.9146 ± 2.7216) (p = 0.1169) \\
\midrule
 External data, POI loss+standard loss &       {\bf 11.2} (11.1627 ± 1.1571)       &        {\bf 31.2} (31.1684 ± 1.2127)        &       90.8 (90.9146 ± 2.7216)        \\
        No external data, POI loss+standard loss & 9.3 (9.2560 ± 1.0709) (p = 0.0040)* & 29.2 (29.1112 ± 1.2121) (p = 0.0020)* & 89.4 (89.4502 ± 2.2439) (p = 0.1299) \\          
\bottomrule
\end{tabular}
\\[.3cm]
\caption{EXP2: {\tt sacrebleu} results for the CAR data for stage 2, comparing the standard Transformer loss function with the POI loss function constraining the generator. Boldface and * indicate significantly better results according to {\tt sacrebleu}.} 
\label{em-sb-s2}
\end{center}
\end{table}

Table \ref{em-nlgeval-s2} displays the {\tt nlg-eval} scores for the conditions of Table \ref{em-sb-s2}.
\begin{table}[H]
\begin{center}
\begin{tabular}{r|p{2cm}|p{2cm}|p{2cm}|p{2cm}}
\toprule
 Metric & External data, standard loss & External data, POI loss + standard loss& No external data, standard loss & No external data, POI loss + standard loss\\
 \midrule
BLEU-1& 34.3 & {\bf 36.0} &  34.3 & 32.3\\
BLEU-2& 22.5 & {\bf 23.8} & 22.4 & 20.7\\
BLEU-3&  14.7 & {\bf 15.9} & 14.9 & 13.8\\
BLEU-4&  9.9 & {\bf 11.2} & 10.3 & 9.3\\
METEOR&  17.5 & {\bf 18.2} & 17.5& 16.6\\
ROUGE-L& 32.6 & {\bf 32.9} & {\bf 32.9} & 31.0\\
SkipThoughtsCosineSimilarity& 53.6 & {\bf 54.2} &  54.0 & 52.7\\
EmbeddingAverageCosineSimilarity & 74.3 & {\bf 74.4} &  74.3 &71.4 \\
VectorExtremaCosineSimilarity & 55.8 & {\bf 55.9} &  55.5 & 52.9\\
GreedyMatchingScore& 76.6 & {\bf 78.0} & 77.2 & 76.0\\
\bottomrule
\end{tabular}
\\[.3cm]
\caption{EXP2: metric results produced by {\tt nlg-eval} for the CAR data for stage 2, comparing the standard Transformer loss function with the POI loss function constraining the generator. Boldface indicates best results.} 
\label{em-nlgeval-s2}
\end{center}
\end{table}

According to {\tt sacrebleu}, using external data with the POI loss function combined with standard loss in stage 2 is significantly better than using external data with just the standard loss function. Further, using the memory buffer in conjunction with POI loss and standard loss significantly outperforms using no external data and POI loss combined with standard loss. This confirms the benefit of the extra memory buffer. 
The results for using no external data, POI loss + standard loss are on a par with using no external data with just the standard loss.

While BLEU scores decrease significantly for stage 2 compared to stages 1 and 3, strict string similarity appears to be not a suitable metric for evaluating this stage. Recall that in stage 2, answer patterns are produced for the attenuated input patterns that are produced by stage 1. A manual inspection for accuracy revealed that the stage 2 patterns produced by Experiment 1 displayed an answer accuracy of 77.7\% with accuracy meaning here: leading to a correct answer to the original question, i.e. having the correct slots, but not necessarily matching formulations.
Some examples are:

{\it Input:} when is poitypetennis activity\\
{\it External data:} poiparty poidate poievent poiagenda poitime\\
{\it Ground truth:} you have a poievent activity on poidate at poitime\\
{\it Prediction:}your poievent activity is on poidate at poitime
\\[.3cm]
{\it Input:}: who is going\\
{\it External data:} poiparty poidate poievent poiagenda poitime\\
{\it Ground truth:} poiparty will be at the poievent activity on poidate at poitime\\
{\it Prediction:} poiparty will be attending your poievent activity on poidate at poitime
\\[.3cm]
{\it Input:} find some places downtown where i can poitype\\
{\it External data:} poiaddress poidistance poitype poi\\
{\it Ground truth:} the poi is poidistance away\\
{\it Prediction:} you will be able to poitype located at poiaddress it is poidistance away\\[.3cm]

{\bf EXP3 - Personalized bAbI}. Table \ref{experiment3} lists the BLEU-4 results for the Personalized bAbI task. While this dataset is small and has repetitive answers, using external data yields significantly better performance. Table \ref{experiment3-metrics} lists the metric results.
\begin{table}[h!]
\begin{center}
\begin{tabular}{rp{3.5cm}p{3.5cm}p{3.5cm}}
\toprule
System & BLEU-4 ($\mu$ ± 95\% CI) & chrF2++++ ($\mu$ ± 95\% CI)& TER ($\mu$ ± 95\% CI)\\
\midrule
 External data &       {\bf 61.1} (60.9185 ± 5.7854)       &        {\bf 67.7} (67.6811 ± 5.0516)        &        {\bf 37.4} (37.3672 ± 5.6361)        \\ \\
        No external data & 8.8 (8.7159 ± 3.1313) (p = 0.0010)* & 20.8 (20.7812 ± 2.6477) (p = 0.0010)* & 97.5 (97.5206 ± 4.4556) (p = 0.0010)* \\
\bottomrule
\end{tabular}
\\[.3cm]
\caption{EXP3: the {\tt sacrebleu} results for the Personalized bAbI data. Boldface and * indicate significantly better results according to {\tt sacrebleu}.} 
\label{experiment3}
\end{center}

\end{table}

\begin{table}[H]
\begin{center}
\begin{tabular}{r|p{3cm}|p{3cm}}
\toprule
 Metric & External data & No external data\\
 \midrule
BLEU-1& {\bf 67.2} & 21.8\\
BLEU-2& {\bf 64.5} & 14.6\\
BLEU-3& {\bf 62.8} & 10.7\\
BLEU-4&  {\bf 61.1} & 8.8\\
METEOR& {\bf 39.5} & 11.1\\
ROUGE-L& {\bf 65.6} & 22.5\\
SkipThoughtsCosineSimilarity& {\bf 78.6} &  46.8\\
EmbeddingAverageCosineSimilarity& {\bf 89.0} & 80.1\\
VectorExtremaCosineSimilarity & {\bf 73.6} & 50.0\\
GreedyMatchingScore& {\bf 82.6} & 63.4\\
\bottomrule
\end{tabular}
\\[.3cm]
\caption{EXP3: metric results produced by {\tt nlg-eval} for the Personalized bAbI data. Boldface indicates best results.} 
\label{experiment3-metrics}
\end{center}
\end{table}


\section{Discussion}

In our experiments, we have found initial indications that informing adversarially trained Transformer models with additional external data and constrained generators can be helpful. Using just external data with the standard loss function appears not to outperform the standard setting where no external data is used. However, we found that using external data with a tailored loss function (POI loss) compared to the standard loss function improves stage 2 of the CAR data experiment.  Using POI loss as a condition on the Transformer generator also produces better absolute results across all metrics minus 1 for stage 2 of the CAR data. The use of the extra memory buffer plus the POI loss function outperformed using no external data and POI loss applied to the input buffer and predictions. This indicates that the extra memory is indeed useful for this task. Our results can be interpreted as a within-system form of ablation, since we tested for using no external data versus using external data, effectively shutting the extra memory buffer off and on.
For Stage 2 of the CAR dataset, we effectively used the extra memory buffer to emphasize the data already in the input buffer - the buffer basically repeats that information. In stage 3 for the CAR dataset, it is clear that the external data (which, in that stage, consists of non-repeated database values) is crucial for filling in the slots names with specific values. For the Personalized bAbI dataset, the situation is actually similar: the external data is crucial for setting the right tone of voice, which is dependent on the age and gender of the human interlocutor. Such data is not redundantly available in the input buffer. The Personalized bAbI task involves only the extra conditioning of answer generation on two external variables (the gender and age of the human interlocutor). This may explain the effectiveness of external data we observed for this dataset.

Limitations of our research consist of a partial assessment of the use of additional loss functions, and the relatively small sample size for our training and test datasets. Also, for practical (computing) reasons, we did not perform a systematic grid search over the various Transformer parameters, including the number of training epochs.  We have not addressed the full potential of conditioning the adversarial aspect of our approach yet. For instance, coding aspects of past utterances in dialogues (like sentiment, style, or topic) may lead to better regulation of future utterances, in conjunction to the general in-context learning capabilities of large language models. Finally, we are aware that our evaluation metrics address only string and semantic overlap between ground truth and predicted data. Follow-up research should extend these metrics with more fine-grained evaluation metrics that target factual adherence to external data.

\section{Future work}\label{futurework}
As for future work, we see a number of potential topics:
\\[.3cm]
{\bf End-to-end training}  In our current setup for factual question-answering, we have trained three memory-augmented Transformers in isolation. An end-to end schema would imply joint optimization with potential performance benefits, and will be investigated in our follow-up research.\\[.3cm] 
{\bf Quality improvement for factual adherence} Additional loss functions will need to be applied in order to tighten the adherence of the GAT models to factual data, as witnessed by the frequent hallucination in stage 3 for the CAR dataset. Factual evaluation metrics will be used for a better estimation of factual adherence.\\[.3cm]
{\bf Structured external data} Structured external data in the form of knowledge graphs or ontologies can improve the answers in stage 2 and 3 for the CAR dataset. Previous work has shown the value of using knowledge graphs as an external knowledge base to question answering with language models \cite{liu2022joint,yang2023chatgpt}. The CAR dataset currently contains a knowledge base per scenario, structured in a triple format \cite{EricKCM17}. These knowledge bases are manually crafted per scenario and have limited information; useful in a limited experiment setting but not scalable for real-life applications. Large graph databases such as Neo4j \cite{noauthororeditorneo4j} provide a large factbase and allow for discovery of patterns and flexible extensions, as \cite{jiang2021research} demonstrate for the medical domain. For follow-up research, it would be interesting to explore the benefits of different types of knowledge graphs as external fact databases.\\[.3cm]
{\bf Comparison with Retrieval-Augmented Generation (RAG) models} In future work, we intend to compare the GAT architecture to RAG architectures, on shared datasets. 
\\[.3cm]
{\bf Reinforcement learning from explicit human feedback}
Reinforcement learning from human feedback (RLHF) \cite{cristiano2017deep, ouyang2022training} can potentially further condition the GAT by providing a mechanism for improving performance based on human-generated rewards or evaluations. This presupposes the detection of appropriate reward signals that capture the desired behavior of the system. For example, in the case of factual question answering, the reward signal could be based on the accuracy of the answer provided by the system compared to a human-provided answer, but does not have to involve a direct human-in-the-loop. A reward model can be trained using human labelers, which in turn can be used to provide a reward signal, making the process time- and cost-efficient. This technique can be applied to all three stages of the dialogue process described above.
Sometimes, based on new evidence or research methods, factual knowledge can change.  By incorporating RLHF, the memory can be updated based on the reward policy. This allows the system to dynamically adapt its memory content and attention focus based on the feedback received through reinforcement learning.\\[.3cm]
{\bf Reinforcement learning from implicit human feedback}
As human-system dialogues become more natural, human feedback to system output is better reflecting the grounding process that takes place in human-human conversations. This means that during a conversation partners are continuously making sure their utterances are mutually understood (\cite{clark96,clark89,clark91}). After the presentation phase, in which the first partner presents an utterance,  the acceptance phase follows, in which the second partner provides evidence of understanding. Grounding is often achieved through the use of back-channel responses, such as “uh huh” or “mm” (\cite{schegloff82}). However, the acceptance phase may take several turns, including sequences of clarification requests and repairs. Once both phases are complete, it will be common ground between both partners that the second partner has understood what the first partner meant.
In addition, this whole process may involve dialog acts indicating the second partner is processing the utterance of the first partner (autofeedback), or the partner needs some time to formulate his contribution the dialog (time management). Problems with understanding or formulating may also become apparent in delays in responding (\cite{bunt2011}).
The response to the grounded utterance of the first partner depends on the dialog act of this utterance and the corresponding expected dialog act of the second partner. These expected pairs of dialog acts are called adjacency pairs, of which typical examples are: question–answer, greeting–greeting and offer–acceptance (\cite{schegloff73}). The response may turn out to adhere to this expectation (e.g., offer-acceptance), contradict it (e.g., offer–refusal) or leave it undecided (e.g. offer-hesitation).
Both length and nature of the process of achieving common ground, occurrence of autofeedback and time management dialog acts or delays in answering, and the adherence of the ultimate response of the human to the system’s dialog act may provide valuable input to reinforcement learning.

The exact interplay of memory augmentation and feedback will be an interesting new research topic.
\\[.3cm]
{\bf Fine-tuning} Our experiments started {\it ab ovo} with a proprietary dataset and an untrained Transformer. In future work, we will attempt to reconcile our approach with pre-trained large language models ({\it base} or {\it foundation} models) by interpreting our memory augmentation as a form of fine-tuning.

\section{Conclusion}
This paper has presented a new type of memory-augmented Transformers: adversarially trained generative Transformers that can be conditioned on arbitrary loss functions imposed on their generators. We have demonstrated the efficacy of an adversarial training scheme for textual Transformers, and found indications that adding external information through memory augmentation leads to performance improvement of our models, for two divergent tasks: factual question-answering and style adaptation.  Our approach is completely agnostic with respect to the type of external data, and may lead to insights which types of external data are beneficiary for a given task at hand. For handling factual question-answering, we outlined an approach based on attenuation and de-attenuation (slot detection, slot mapping, and slot filling), using memory augmentation for emphasizing certain parts of the input data.  While we did not arrive at high quality factual adherence in the slot filling stage of these latter experiments yet, we hypothesize that additional value-checking loss functions will be effective for raising performance. Memory augmentation with truly external data (i.e. data that is not explicitly repeated in the input buffer) appeared useful for a small scale style adaptation dataset. While our results are based on limited data and therefore should not be taken as conclusive, we feel encouraged to further explore the proposed approach across different tasks and on larger datasets.


\section*{Acknowledgments}
This research was partially supported by the TNO Research Programme APPL.AI (https://appl-ai-tno.nl).

\bibliographystyle{unsrt}  
\bibliography{references}

\end{document}